\documentclass[10pt, a4paper]{article}

\usepackage{lrec2026} 

\usepackage{times}
\usepackage{latexsym}

\usepackage[utf8]{inputenc}
\usepackage[T5]{fontenc} 
\usepackage[vietnamese,english]{babel}
\usepackage{multirow}

\usepackage{microtype}
\microtypesetup{protrusion=false}

\usepackage{comment}
\usepackage{graphicx}

\title{FreeTxt-Vi: A Benchmarked Vietnamese–English Toolkit for Segmentation, Sentiment, and Summarisation}
\name{
Hung Nguyen Huy\textsuperscript{1},
Mo El-Haj\textsuperscript{1,3},
Dawn Knight\textsuperscript{2},
Paul Rayson\textsuperscript{1,3}
}

\address{
\textsuperscript{1}VinUniversity, Vietnam\\
\textsuperscript{2}Cardiff University, UK\\
\textsuperscript{3}Lancaster University, UK\\
{\small \texttt{25hung.nh@vinuni.edu.vn, elhaj.m@vinuni.edu.vn}}\\
{\small \texttt{p.rayson@lancaster.ac.uk, knightd5@cardiff.ac.uk}}
}

\abstract{FreeTxt-Vi is a free and open-source web-based toolkit for creating and analysing bilingual Vietnamese–English text collections. Positioned at the intersection of corpus linguistics and natural language processing (NLP), it enables users to build, explore, and interpret free-text data without requiring programming expertise. The system combines established corpus analysis features such as concordancing, keyword analysis, word relation exploration, and interactive visualisation with modern transformer-based NLP components for sentiment analysis and summarisation. A key contribution of this work is the design of a unified bilingual NLP pipeline that integrates a hybrid VnCoreNLP + Byte Pair Encoding (BPE) segmentation strategy, a fine-tuned TabularisAI sentiment classifier, and a fine-tuned Qwen2.5 model for abstractive summarisation. Unlike existing text analysis platforms, FreeTxt-Vi is evaluated as a set of language processing components. We conduct a three-part evaluation covering segmentation, sentiment analysis, and summarisation, and demonstrate that our approach achieves competitive or superior performance compared to widely used baselines in both Vietnamese and English. By reducing technical barriers to multilingual text analysis, FreeTxt-Vi supports reproducible research and promotes the development of language resources for Vietnamese, a widely spoken but underrepresented language in NLP. The toolkit is applicable to a wide range of domains, including education, digital humanities, cultural heritage, and the social sciences, where qualitative text data are common but often difficult to process at scale.
 \\ \newline \Keywords{Vietnamese, English, bilingual corpora, free-text processing, language resources, sentiment analysis, summarisation, NLP evaluation} }

\begin{document}

\maketitleabstract

\section{Introduction}

Free-text qualitative data collected through surveys, questionnaires, and feedback forms carry valuable insights into human experiences and opinions. However, analysing such unstructured text at scale remains challenging, particularly in multilingual and low-resource settings where accessible tools are scarce. While high-resource languages such as English benefit from mature NLP pipelines and pretrained models \cite{devlin-etal-2019-bert}, many languages continue to be underrepresented in language technology research \cite{joshi2021statefatelinguisticdiversity}.

Vietnamese, spoken by more than 80 million people, exemplifies this challenge. Despite recent progress from the Vietnamese NLP community, including resources such as VnCoreNLP \cite{vu2018vncorenlp}, VLSP shared tasks \cite{nguyen2018vlsp}, and pretrained models such as PhoBERT \cite{nguyen2020phobert} and ViT5 \cite{phan2022vit5}, most available tools remain difficult to deploy without programming expertise. Consequently, there is a gap between advances in Vietnamese NLP and their practical uptake in domains such as education, healthcare, cultural heritage, and social research.

Existing text analysis platforms such as Voyant Tools \cite{sinclair2016voyant} and FreeTxt \cite{KNIGHT2024100103} demonstrate the value of combining NLP with interactive corpus analysis. However, these systems do not support Vietnamese and do not address the specific linguistic challenges associated with word segmentation in Vietnamese or bilingual analysis \cite{khallaf2025freetxt}. To date, no open system offers practical, bilingual Vietnamese--English free-text processing with evaluated NLP components.

To address this gap, we introduce FreeTxt-Vi, an open-source web-based toolkit for Vietnamese and English free-text analysis. The system integrates corpus linguistic methods with modern NLP functionality, including sentiment analysis, extractive and abstractive summarisation, concordancing, keyword analysis, and word relation visualisation. It uses a bilingual processing pipeline with a hybrid VnCoreNLP \cite{vu2018vncorenlp} and Byte Pair Encoding (BPE) \cite{sennrich-2016-bpe} segmenter, a fine-tuned TabularisAI model for sentiment analysis \cite{tabularisai_2025}, and a fine-tuned Qwen2.5 model for summarisation \cite{qwen2.5}.

This paper makes three contributions. First, it presents the architecture and implementation of FreeTxt-Vi as a practical system for Vietnamese--English text processing. Second, it introduces and releases the hybrid segmentation pipeline, fine-tuned sentiment analysis model, and bilingual summarisation component, extending existing Vietnamese NLP resources. Third, it provides a three-part evaluation covering segmentation, sentiment analysis, and summarisation, demonstrating that FreeTxt-Vi achieves competitive or superior performance compared to widely used baselines. All tools, resources, and implementation are made freely available for research use\footnote{\url{https://github.com/VinNLP/Free-txt-vi}}, supporting transparency and reproducibility. By lowering technical barriers and providing an evaluated bilingual processing pipeline, FreeTxt-Vi contributes practical infrastructure for Vietnamese NLP and multilingual text analysis.

\section{Related Work}

\subsection{Corpus and Text Analysis Tools}

The analysis of qualitative free-text data has long been supported by digital tools developed in both the social sciences and linguistics. Computer Assisted Qualitative Data Analysis Software (CAQDAS) such as NVivo\footnote{\url{https://lumivero.com/products/nvivo/}} and ATLAS.ti\footnote{\url{https://atlasti.com}} enable users to code, annotate, and manage large volumes of qualitative data. These platforms are widely used in disciplines such as education, sociology, and marketing, but they typically emphasise manual coding frameworks over automated computational analysis. While they provide valuable organisational capabilities, their automated linguistic analysis is often limited and restricted to English or other major languages.

In contrast, tools from corpus linguistics offer automated approaches to text analysis. AntConc \cite{anthony2019antconc}, a widely used freeware concordancer, supports keyword extraction, collocation analysis, and concordance generation, providing linguists with detailed frequency-based analyses. Similarly, Sketch Engine \cite{kilgarriff2014sketchengine} offers large-scale corpus management and advanced queries, though access is subscription-based and its usability for non-specialists is limited. Wmatrix \cite{rayson2004wmatrix} extends these approaches by integrating semantic annotation and keyness analysis, enabling comparisons across corpora. Although these tools provide powerful functionality, they often require specialist training and are not tailored to survey-style, bilingual data.

Recent efforts have also sought to make text analysis more accessible to non-technical audiences. Voyant Tools \cite{sinclair2016voyant} is a popular web-based platform that allows users to perform frequency counts, collocation analysis, and word cloud visualisations through a browser interface. While highly accessible, its analytical depth is restricted and multilingual support is limited. Similarly, LIWC (Linguistic Inquiry and Word Count) \cite{pennebaker2015development} offers psychologically motivated text categories but focuses primarily on English and has limited flexibility in adapting to low-resource contexts.

FreeTxt \cite{KNIGHT2024100103} was introduced to address these limitations, particularly for minority language contexts. Initially designed to support English and Welsh, it integrated established corpus methods (keyword in context, concordances, frequency lists) with newer NLP components such as sentiment analysis and summarisation. By providing an accessible web-based interface, FreeTxt demonstrated how corpus and NLP techniques could be packaged for non-specialists, particularly in cultural heritage organisations in Wales. FreeTxt-Vi builds directly on this foundation while continuing the commitment to supporting low-resource languages in new contexts. Whereas the original FreeTxt concentrated on a European minority language (Welsh), FreeTxt-Vi extends this vision to Vietnamese, a global low-resource language with distinct challenges such as word segmentation and resource scarcity. It integrates Vietnamese-specific NLP tools, notably VnCoreNLP, to handle compound words and idiomatic expressions effectively. In addition, FreeTxt-Vi broadens the functionality by incorporating abstractive summarisation with large language models and enhanced visualisations tailored to bilingual Vietnamese–English datasets. Most importantly, this work demonstrates how open-source platforms can be adapted and reconfigured to accommodate the linguistic characteristics of new languages, showing a clear pathway for extending accessible NLP resources to other under-served language communities.

\subsection{Vietnamese NLP Resources}
Although Vietnamese is spoken by more than 80 million people worldwide, it has long been underrepresented in NLP research compared to languages such as English, Chinese, and French. A major linguistic challenge lies in Vietnamese word segmentation: unlike alphabetic languages, Vietnamese orthography uses spaces to separate syllables rather than words, meaning that multi-syllabic words frequently appear as separate tokens. This creates downstream challenges for syntactic parsing, NER, and language modelling. Despite growing interest in Southeast Asian languages, publications focusing on Vietnamese in major ACL Anthology\footnote{\url{https://aclanthology.org/}}
 venues remain limited, with typically fewer than five papers per year addressing Vietnamese NLP, highlighting its continued underrepresentation.

Important progress has nonetheless been made within the Vietnamese NLP community. \citet{vu2018vncorenlp} introduced VnCoreNLP, a widely adopted toolkit that provides word segmentation, part-of-speech tagging, named entity recognition, and dependency parsing. It has since become the de facto standard for Vietnamese preprocessing and forms an integral component of the FreeTxt-Vi pipeline. In parallel, community-driven efforts such as the VLSP (Vietnamese Language and Speech Processing) workshops have advanced the development of benchmark datasets and evaluation standards. VLSP shared tasks have focused on core NLP problems including word segmentation, named entity recognition, and machine translation \cite{nguyen2018vlsp}, contributing significantly to resource creation and scholarly engagement.

Building on these foundations, Vietnamese NLP has entered the era of large-scale pretrained language models. PhoBERT \cite{nguyen2020phobert}, an adaptation of the RoBERTa architecture for Vietnamese, was trained on billions of Vietnamese words and achieved state-of-the-art performance across a range of NLP tasks. More recently, ViT5 \cite{phan2022vit5} extended the T5 sequence-to-sequence architecture to Vietnamese, producing strong results in text generation and summarisation. Resources for bilingual and cross-lingual research have also expanded. Parallel corpora such as the IWSLT 2015 dataset \cite{cettolo2016iwslt} and EVBCorpus \cite{ngo2013evbcorpus} provide extensive English–Vietnamese sentence-aligned data, enabling advances in machine translation and multilingual modelling. However, these corpora are primarily geared towards translation tasks and offer limited applicability for qualitative or domain-specific text analysis.

Despite these developments, Vietnamese NLP remains largely inaccessible to non-specialists. Many available tools—such as VnCoreNLP, PhoBERT, and ViT5—require programming proficiency and familiarity with machine learning frameworks. While general-purpose corpus analysis platforms like AntConc, Sketch Engine, and Voyant Tools offer partial support for Vietnamese, they lack robust language-specific processing, especially with respect to segmentation and named entity recognition. Consequently, there is a notable absence of user-friendly platforms that allow researchers, educators, and professionals outside computer science to analyse Vietnamese text effectively. Addressing this gap is the primary motivation behind FreeTxt-Vi, which aims to leverage existing NLP resources while providing an accessible interface for practical Vietnamese language analysis.

Beyond Vietnamese, several platforms have attempted to support multilingual or bilingual text analysis. Tools such as LancsBox\footnote{\url{https://lancsbox.lancs.ac.uk/}} integrate concordance and collocation analysis with graphical visualisations, and support multiple languages through Unicode compatibility. However, language-specific resources (e.g., lemmatisers, taggers) are not consistently available for under-resourced languages. Similarly, commercial platforms such as Google Cloud NLP or IBM Watson offer multilingual text analytics APIs, but they are not open-source, can be costly, and often lack fine-grained control over linguistic processing.

In the context of minority and low-resource languages, FreeTxt \cite{KNIGHT2024100103} demonstrated the value of co-design with stakeholders. Its bilingual English–Welsh functionality showed that cultural and institutional needs could directly shape tool development, particularly in contexts where survey respondents are legally entitled to answer in either language. FreeTxt-Vi follows this approach but targets a new linguistic context. Unlike Welsh, which benefits from government support and corpus projects such as CorCenCC \cite{corcencc}, Vietnamese NLP has been largely community-driven and remains relatively fragmented. The introduction of an accessible, bilingual tool that leverages VnCoreNLP and pretrained Vietnamese models provides both continuity with FreeTxt’s original vision and a novel contribution to Vietnamese language technology.

\section{Design and Implementation}
\label{sec:designAndImplementation}

\subsection{Data Uploading and Preprocessing}

FreeTxt-Vi is designed to support a wide range of data formats commonly used in qualitative and survey-based research. Users may upload plain text (\texttt{.txt}), spreadsheets (\texttt{.xlsx}), or comma-separated values (\texttt{.csv}) files, and the platform also provides a direct text input option to enable rapid exploratory analysis without prior file preparation.

Following upload, the system automatically identifies the dominant language (English or Vietnamese) using a lightweight language detection module. This step is essential for correctly routing mixed-language datasets, which are common in bilingual research settings, to the appropriate preprocessing pipeline.

Vietnamese text is processed using VnCoreNLP \cite{vu2018vncorenlp}, which performs word segmentation, part-of-speech tagging, and named entity recognition. Accurate segmentation is particularly important for Vietnamese since words often consist of multiple whitespace-separated syllables (e.g.\ \textit{học sinh} “student”); improper segmentation can distort frequency counts, collocation extraction, and downstream visualisations. English preprocessing relies on standard NLP libraries (NLTK and spaCy) for tokenisation and sentence splitting. This dual-language preprocessing layer ensures consistent and linguistically appropriate handling of input data, forming a reliable basis for subsequent analytical modules.

\subsection{Sentiment Analysis}

Sentiment analysis is implemented as a core analytical component in FreeTxt-Vi to support rapid assessment of emotional or attitudinal content in text data. The system employs a multilingual transformer-based model fine-tuned for sentiment classification \cite{tabularisai_2025}, enabling analysis of both Vietnamese and English datasets within the same workflow.

\begin{figure}[!ht]
\centering
\includegraphics[width=0.8\columnwidth]{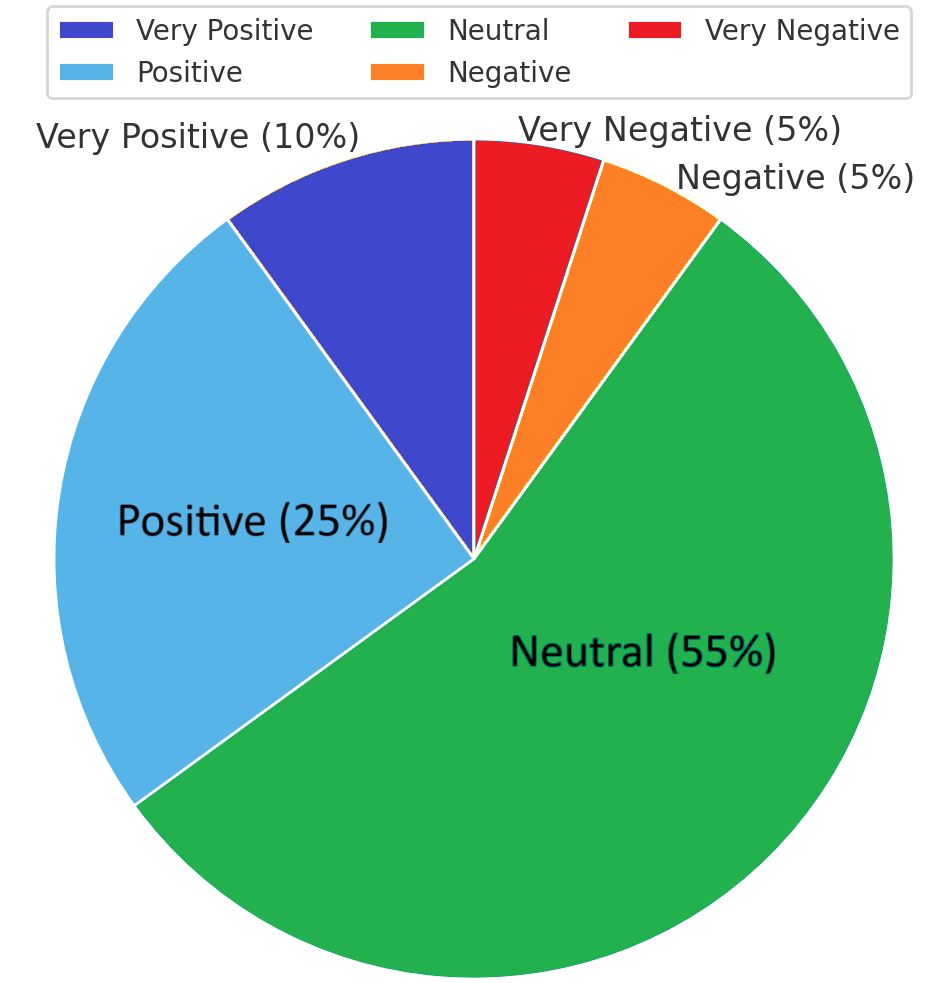}
\caption{Example output of FreeTxt-Vi’s sentiment analysis module showing five-class classification.}
\label{fig:sentiment_analysis}
\end{figure}

\begin{figure}[!ht]
\begin{center}
\includegraphics[width=\columnwidth]{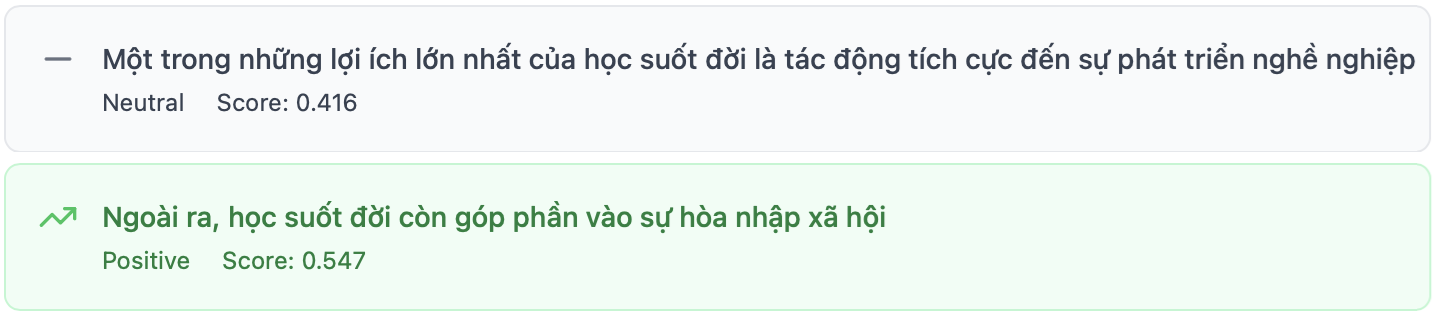}
\caption{Sentiment output example, based on 5 class sentiment.}
\label{fig.2}
\end{center}
\end{figure}

Users may choose between coarse-grained sentiment labels (positive, negative, neutral) or a more detailed five-level output (very positive, positive, neutral, negative, very negative), depending on their analytical requirements. Outputs are displayed through interactive visual interfaces, including sentiment distribution bar and pie charts and ranked sentiment tables with confidence scores. Examples are shown in Figures~\ref{fig:sentiment_analysis} and~\ref{fig.2}.

For Vietnamese input, sentiment analysis is enhanced by prior VnCoreNLP segmentation, which ensures that sentiment-bearing multiword expressions are preserved during classification. Since sentiment in Vietnamese is frequently expressed through phrasal units, segmentation-aware processing is essential for classification accuracy. By integrating transformer-based modelling with language-specific preprocessing, FreeTxt-Vi delivers sentiment analysis that is both scalable and sensitive to linguistic variation across English and Vietnamese.

\subsection{Summarisation}
Survey responses and open-ended feedback are often long and heterogeneous, making them difficult to process manually. FreeTxt-Vi addresses this challenge by providing two complementary summarisation approaches.

The first is an extractive method based on the TextRank algorithm \cite{mihalcea-tarau-2004-textrank}, which identifies the most salient sentences by modelling the text as a graph and ranking sentences according to their centrality. Extractive summarisation is efficient and transparent: users can see precisely which sentences were chosen and why.

The second summarisation option in FreeTxt-Vi is an abstractive module powered by the Qwen 2.5 large language model (LLM) \cite{qwen2.5}. Unlike extractive techniques, which merely select existing sentences, this module is capable of generating concise, fluent summaries that paraphrase and reorganise information, mirroring human reasoning. Crucially, FreeTxt-Vi extends standard summarisation by supporting prompt-guided summarisation, allowing users to steer the model toward analytical or thematic objectives (e.g. “summarise viewpoints on environmental sustainability” or “highlight challenges faced by teachers”). This makes FreeTxt-Vi not just a summarisation tool, but a flexible exploratory analysis system, enabling domain-specific insights from qualitative datasets—functionality not available in existing Vietnamese NLP platforms.

\begin{figure}[!ht]
\begin{center}
\includegraphics[width=\columnwidth]{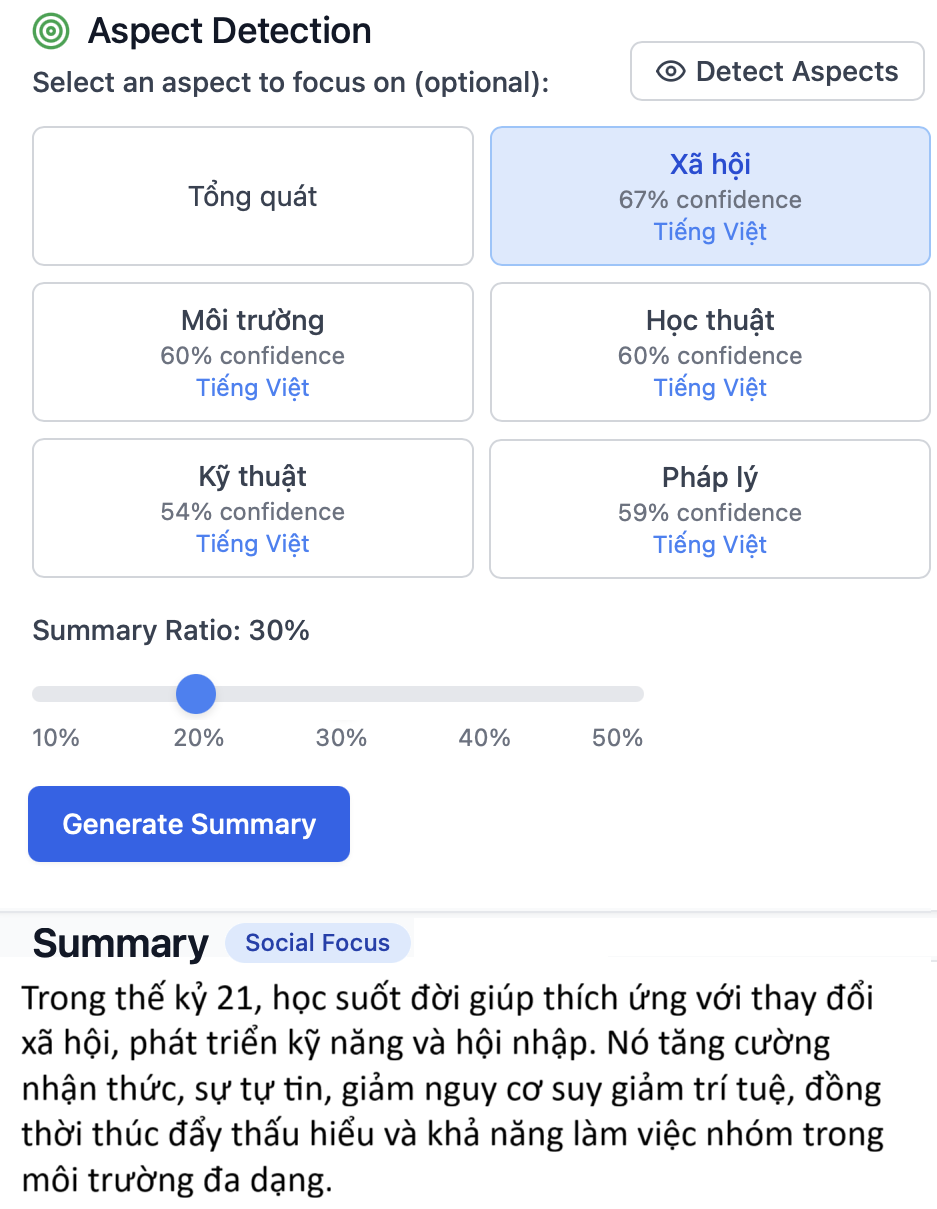}
\caption{FreeTxt-Vi Aspect-based Summarisation.}
\label{fig.3}
\end{center}
\end{figure}

A further innovation in FreeTxt-Vi is its aspect-guided abstractive summarisation, which extends the Qwen 2.5 LLM beyond general summarisation to targeted thematic analysis (Figure~\ref{fig.3}). The system automatically detects salient aspects in the data—such as social, academic, environmental, or technical themes—and assigns confidence scores. Users can then select a specific aspect (e.g. Xã hội / Social) to generate a focused summary that foregrounds only the content relevant to that dimension. This enables fine-grained exploration of qualitative feedback, such as isolating social concerns in educational surveys or extracting customer pain points from service reviews. This capability is novel in the context of Vietnamese NLP, where existing tools lack support for interactive, end-user-driven abstractive summarisation. By combining extractive, free-form abstractive, and aspect-guided summarisation within a single interface, FreeTxt-Vi offers a uniquely flexible and interpretable approach to large-scale Vietnamese–English text analysis, bridging a gap previously dominated by English-centric NLP platforms.

\subsection{Word Cloud}  
Word clouds provide an immediate and intuitive overview of the most salient terms in a dataset. FreeTxt-Vi extends this familiar visualisation by supporting three complementary methods, illustrated in Figure~\ref{fig:vi_wordclouds}:

\begin{figure*}[t]
  \centering
  \begin{minipage}[t]{0.32\linewidth}
    \centering
    \includegraphics[width=\linewidth]{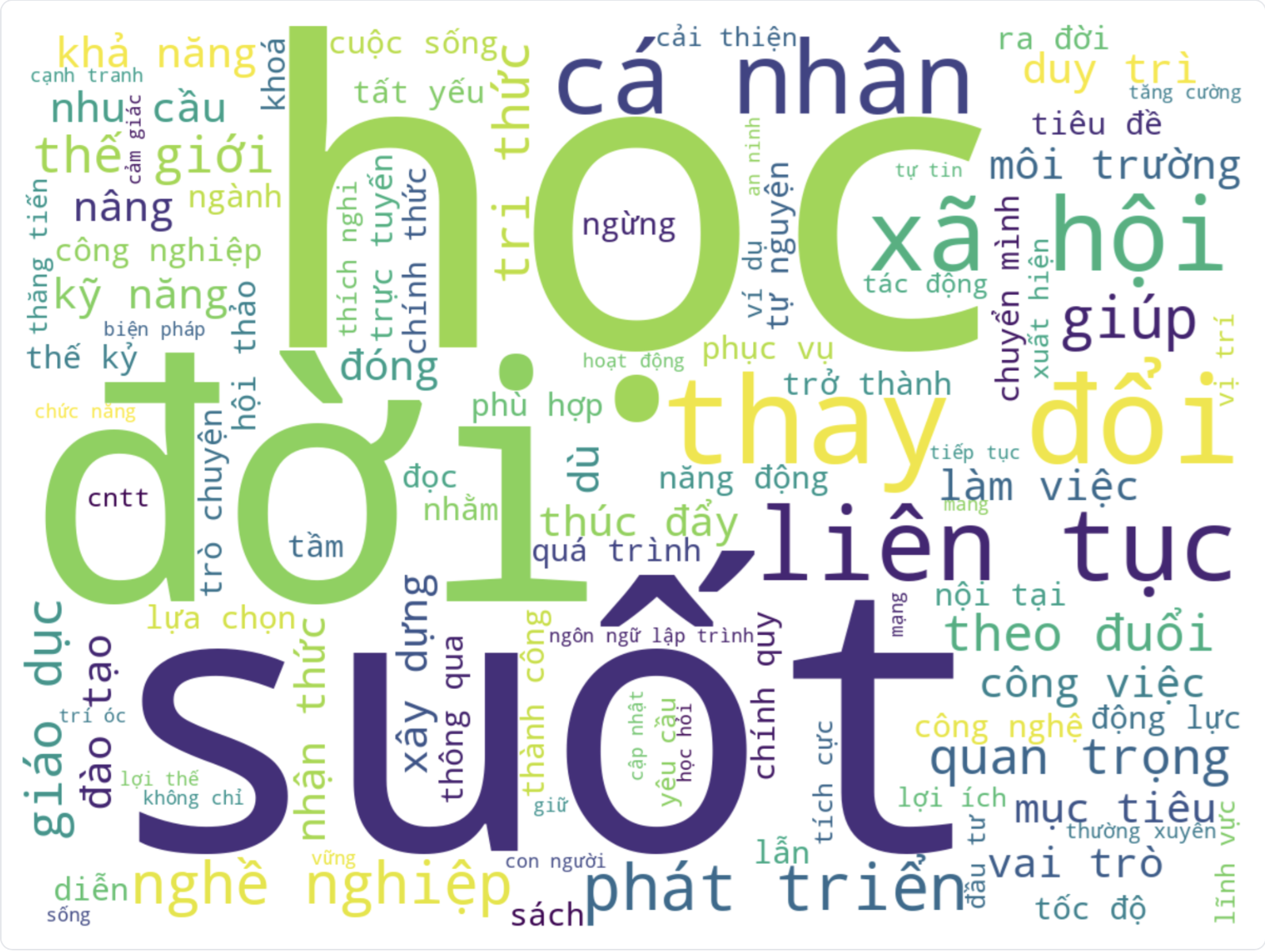}
    \caption{(a) Frequency-based word cloud}
  \end{minipage}\hfill
  \begin{minipage}[t]{0.32\linewidth}
    \centering
    \includegraphics[width=\linewidth]{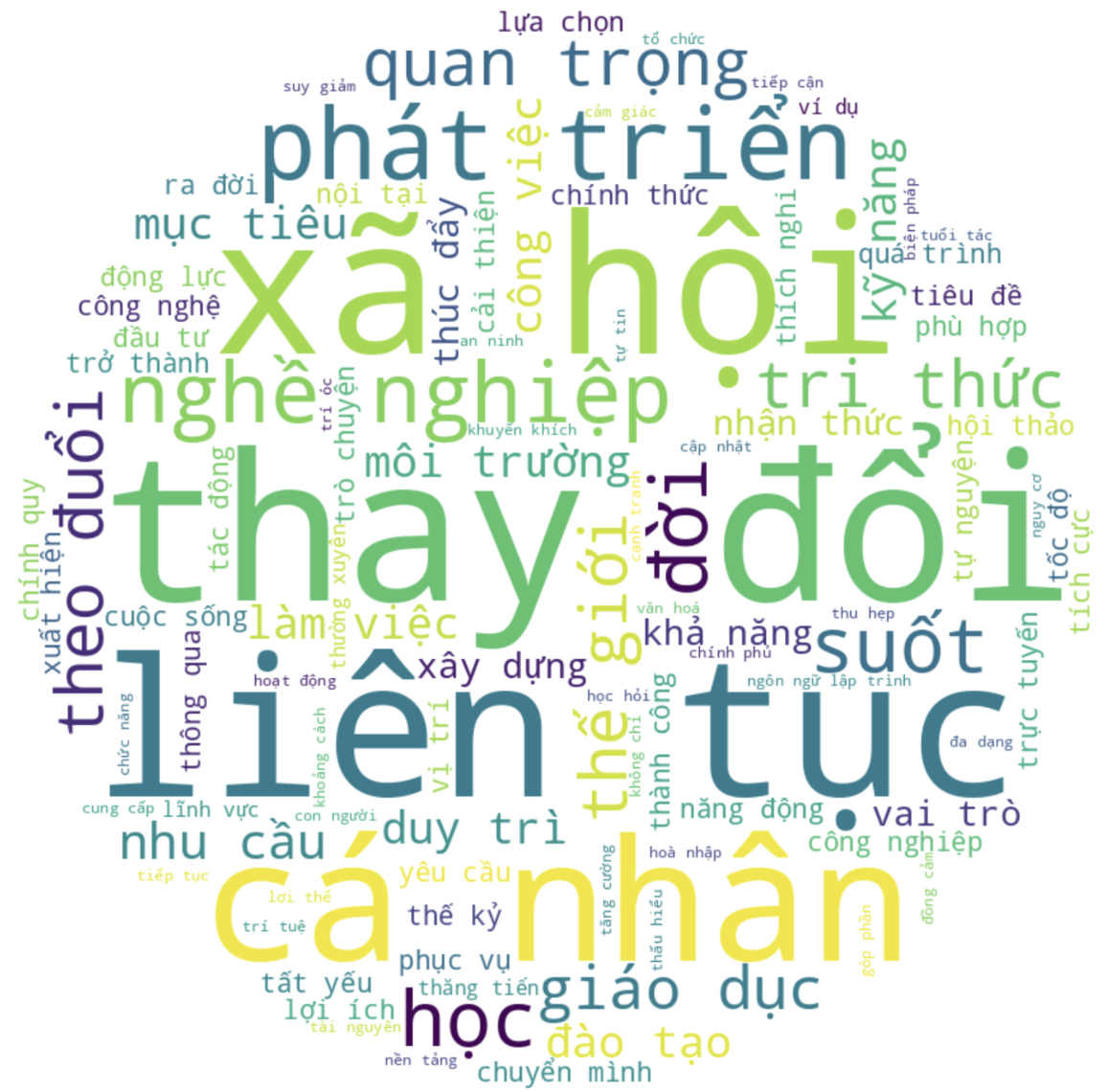}
    \caption{(b) Log-likelihood word cloud}
  \end{minipage}\hfill
  \begin{minipage}[t]{0.32\linewidth}
    \centering
    \includegraphics[width=\linewidth]{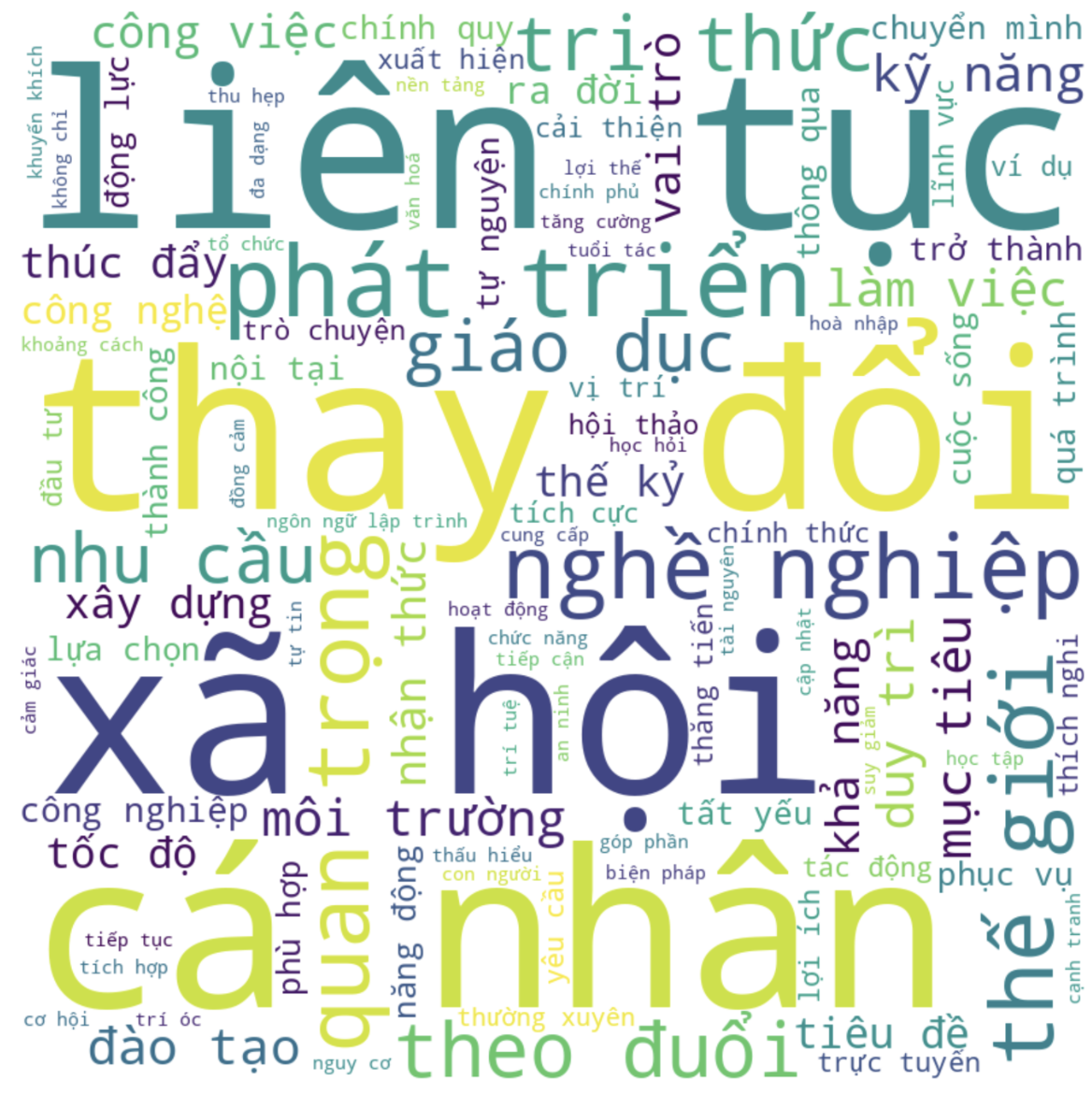}
    \caption{(c) Keyness word cloud}
  \end{minipage}
  \caption{Word cloud visualisations in FreeTxt-Vi for Vietnamese text: (a) frequency-based, (b) log-likelihood-based, and (c) keyness-based. These provide complementary perspectives on word prominence in the dataset.}
  \label{fig:vi_wordclouds}
\end{figure*}

\begin{itemize}
    \item \textbf{Frequency-based word clouds} (Figure~\ref{fig:vi_wordclouds}a) display words according to their raw counts, giving a quick snapshot of dominant vocabulary.  

    \item \textbf{Log-likelihood word clouds} (Figure~\ref{fig:vi_wordclouds}b) highlight terms that occur significantly more often than expected relative to a reference corpus.  

    \item \textbf{Keyness word clouds} (Figure~\ref{fig:vi_wordclouds}c) identify words that are particularly distinctive of the dataset compared with general language use.  
\end{itemize}  

For English data, the British National Corpus (BNC) serves as the baseline, while for Vietnamese, FreeTxt-Vi draws on a frequency dictionary derived from the VietNews dataset \cite{nguyen2019vnds}. This ensures that statistical comparisons are grounded in reliable, language-specific references.  

The system also allows users to remove stopwords, customise scaling, and export word clouds for reporting. By combining raw frequency, statistical distinctiveness, and keyness measures, FreeTxt-Vi delivers a multi-layered view of word prominence. Importantly, this is the first tool to offer such advanced, corpus-informed word cloud visualisations for Vietnamese, demonstrating how open-source platforms can bring established corpus linguistic methods into the analysis of low-resource languages.  

\subsection{Word Tree}
The word tree visualisation, adapted from the original FreeTxt framework \cite{KNIGHT2024100103}, provides an interactive way to explore keyword-in-context (KWIC) patterns. Users can select any keyword, and the system generates branching sequences of words that occur before and after it in the corpus. This allows users to trace recurring expressions, collocational patterns, and discourse structures in an intuitive graphical form.  

For Vietnamese, FreeTxt-Vi integrates VnCoreNLP preprocessing to ensure that multiword expressions are recognised as single tokens rather than fragmented syllables. This is a critical adaptation for Vietnamese, where compounds such as \textit{``công nghệ thông tin''} (information technology) would otherwise be split incorrectly. By preserving these multiword units, the word tree displays meaningful linguistic structures rather than disconnected syllables.  

Users can interact with the tree by expanding or collapsing branches, adjusting the size of the context window, and exporting the results. This makes the tool especially valuable for qualitative discourse analysis, thematic exploration, and stylistic studies, where understanding co-text is essential.  

\begin{figure}[h]
\centering
\includegraphics[width=\columnwidth]{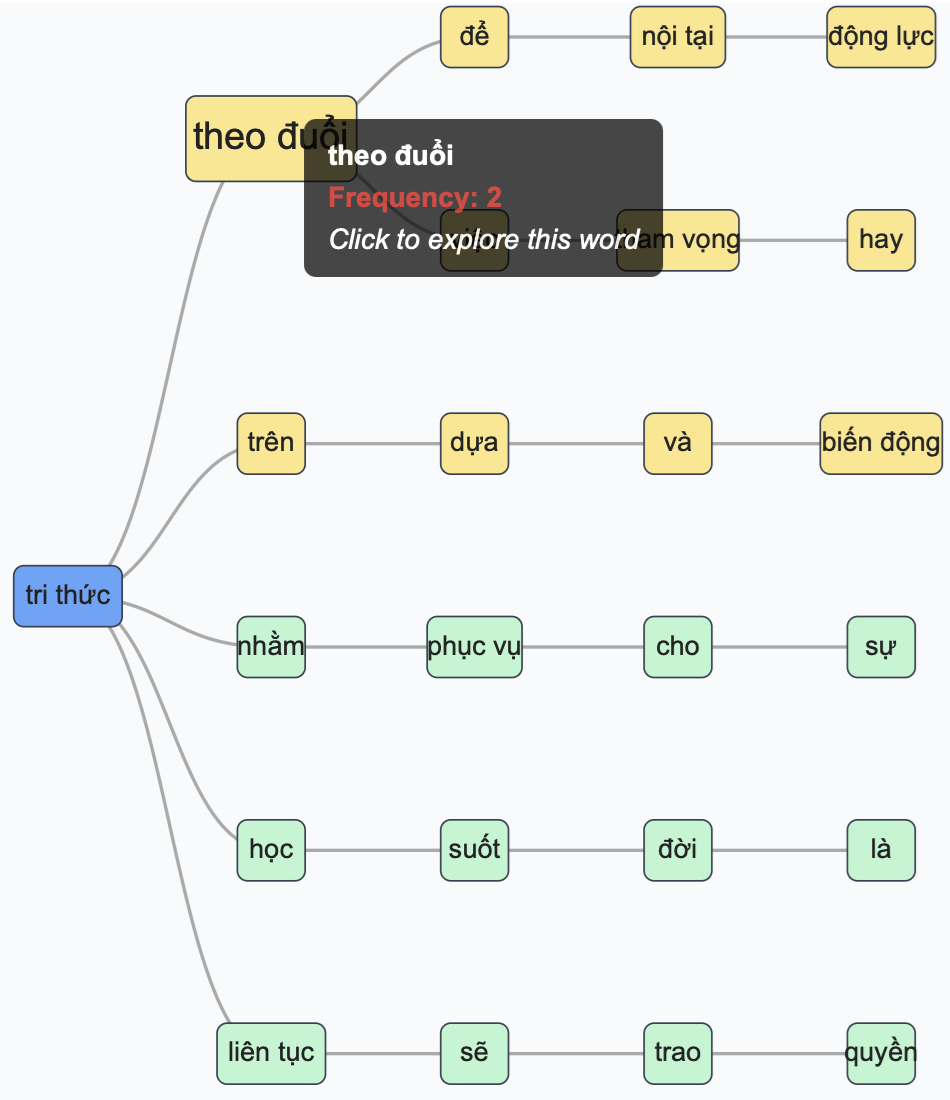}
\caption{Word tree visualisation in FreeTxt-Vi showing the keywords ``tri thức'' (knowledge) and ``theo đuổi'' (pursue) in their co-text.}
\label{fig:vi_word_tree}
\end{figure}

Figure~\ref{fig:vi_word_tree} illustrates this functionality using the keywords \textit{``tri thức''} (knowledge) and \textit{``theo đuổi''} (pursue). The branches show how these terms occur within different contexts, enabling users to explore not just frequency but also the diversity of surrounding linguistic environments.  

Importantly, this is the first implementation of interactive word tree visualisation for Vietnamese text, demonstrating how FreeTxt-Vi extends the capabilities of FreeTxt beyond English and Welsh to a global low-resource language.  

\subsection{Word Concordance}
The concordance tool is a cornerstone of corpus linguistics and remains one of the most widely used methods for qualitative textual analysis. In FreeTxt-Vi, the concordance module displays all occurrences of a chosen keyword together with a configurable window of surrounding context. This enables users to examine how terms are deployed across different textual environments, supporting stylistic, semantic, and discourse-level investigations.  

For Vietnamese text, segmentation is handled through \textbf{VnCoreNLP}, which ensures that keywords and their contexts are identified correctly even when they involve multiword expressions. This adaptation is essential for Vietnamese, where syllable-based orthography would otherwise fragment meaningful tokens (e.g., ``học tập'' for ``study/learning'').  

\begin{figure*}[t]
\centering
\includegraphics[width=0.9\textwidth]{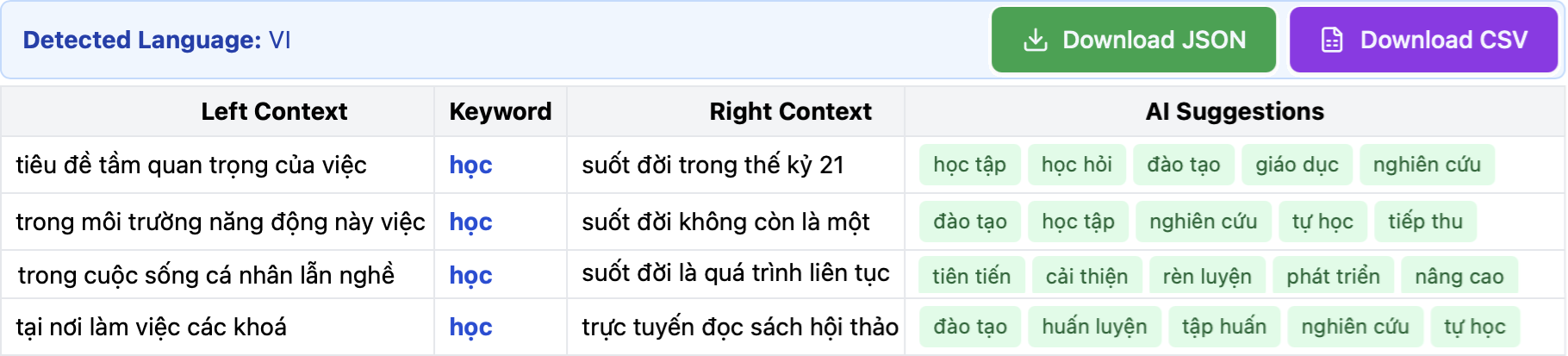}
\caption{Concordance display for the keyword ``học'' (study) in FreeTxt-Vi, showing surrounding context and AI-based lexical suggestions.}
\label{fig:vi_word_concord}
\end{figure*}

To extend the usefulness of concordances, FreeTxt-Vi introduces an optional suggestion module powered by LLMs. In addition to the keyword-in-context (KWIC) display, the system proposes semantically related alternatives—such as synonyms or paraphrased expressions—allowing users to explore lexical variation and discover related concepts. Figure~\ref{fig:vi_word_concord} illustrates this with the keyword \textit{``học''} (study), showing left and right contexts alongside AI-generated suggestions including \textit{``đào tạo''} (training), \textit{``giáo dục''} (education), and \textit{``nghiên cứu''} (research).  

This integration of concordance analysis with AI-driven lexical suggestion is a \textbf{novel feature for Vietnamese text analysis}. While traditional concordancers such as AntConc display KWIC lines, FreeTxt-Vi uniquely combines this close-reading functionality with exploratory flexibility, helping users to investigate both the immediate contexts of words and the broader semantic field in which they operate.  

\section{Evaluation}\label{sec:evaluation}

\subsection{Experimental Setup}

\paragraph{Model Scale.}
TabularisAI multilingual model contains 110M parameters (base architecture).
Qwen2.5-7B contains 7 billion parameters.

\paragraph{Fine-tuning Details.}
Sentiment fine-tuning was conducted for 3 epochs using AdamW (learning rate 2e-5, batch size 32).
Summarisation fine-tuning used LoRA adaptation with rank 8 and learning rate 1e-5 for 2 epochs.

\paragraph{Dataset Sizes.}
VLSP2013 (segmentation): 10,000 sentences (training: 8k, test: 2k).
VLSP2016 (sentiment): 16,000 reviews.
IMDb: 50,000 reviews.
VNDS: 300,000 article-summary pairs.
CNN/DailyMail: 312,000 article-summary pairs.

We conduct an evaluation across three major tasks: word segmentation/tokenization, sentiment analysis, and Summarisation.  
All experiments are implemented in Python, using HuggingFace Transformers and PyTorch backends.  
For Vietnamese, we use the VLSP-2013 dataset for segmentation, VLSP-2016\footnote{\url{https://vlsp.org.vn/}} for sentiment classification, and VNDS \cite{nguyen2019vnds} for Summarisation.  
For English, we use the Wikipedia/OpenWebText dataset for tokenization, IMDb for sentiment analysis, and CNN/DailyMail for Summarisation.  
Evaluation metrics include F1 score for segmentation quality, classification accuracy for sentiment analysis, ROUGE scores for summarisation quality, and processing speed (throughput), reported as sentences, documents, or tokens per second depending on the task.

\subsection{Word Segmentation and Tokenization}

Although BPE, WordPiece, and SentencePiece are not traditional linguistic word segmenters, they are included as baselines because they represent the dominant tokenisation strategies used in transformer architectures. Comparing against them allows us to evaluate whether linguistically informed segmentation provides measurable advantages over purely subword-based tokenisation in Vietnamese.

For Vietnamese and English text processing, we compare our hybrid segmentation and tokenisation approach against a set of widely used baseline tools. For Vietnamese, we include VnCoreNLP \cite{vu2018vncorenlp}, a neural-based pipeline optimised for high-speed linguistic processing, and the NlpHUST word segmenter\footnote{\url{https://huggingface.co/NlpHUST/vi-word-segmentation}}, which was one of the top-performing systems in the VLSP 2013 shared tasks. We also evaluate Underthesea\footnote{\url{https://github.com/undertheseanlp/underthesea}}, a hybrid dictionary-based Vietnamese NLP toolkit. For multilingual and transformer-based pipelines, we compare against subword-based segmentation methods including Byte Pair Encoding (BPE) \cite{sennrich-2016-bpe}, WordPiece \cite{wordpiece2020}, and SentencePiece \cite{sentencepiece2018}, all of which are used in modern pretrained language models. We additionally include BlingFire\footnote{\url{https://pypi.org/project/blingfire/}}, a fast rule-based tokenizer developed by Microsoft for large-scale text processing.

Table~\ref{tab:merged_benchmark_f1_speed_filled} reports results for Vietnamese word segmentation and English tokenisation. Our hybrid method (VnCoreNLP + BPE) achieves strong segmentation performance (98.1\% F1) while maintaining high efficiency (4,900 sentences/sec). Although the Vietnamese-focused NlpHUST segmenter attains a slightly higher F1 score (98.3\%) and standalone VnCoreNLP performs competitively (97.8\%), these tools are not designed for integration into a unified bilingual processing pipeline. By contrast, our method provides a consistent segmentation–tokenisation layer for both languages and incorporates subword modelling through BPE, improving robustness to rare and out-of-vocabulary terms while remaining compatible with transformer architectures. This bilingual advantage is reflected in the English benchmark, where our pipeline achieves 99.94\% tokenisation accuracy and the highest throughput among linguistically informed baselines (75,500 docs/sec). Overall, our approach is the strongest bilingual method, delivering competitive accuracy in Vietnamese while also performing efficiently and reliably for English.

\begin{table}[h]
\centering
\footnotesize
\renewcommand{\arraystretch}{1.2}
\resizebox{\columnwidth}{!}{%
\begin{tabular}{lcc|cc}
\hline
\multirow{2}{*}{\textbf{Tokenizer/Segmenter}} 
  & \multicolumn{2}{c|}{\textbf{VLSP2013 (Vietnamese)}} 
  & \multicolumn{2}{c}{\textbf{Wikipedia/OpenWebText (English)}} \\
\cline{2-5}
  & \textbf{F1 Score (\%)} & \textbf{Speed (sent/sec)} 
  & \textbf{F1/Accuracy (\%)} & \textbf{Speed (docs/sec)} \\
\hline
\textbf{Our (VnCoreNLP + BPE)} & \textbf{98.1} & 4,900 & \textbf{99.94} & \textbf{75,500} \\
WordPiece (BERT) & 91.2 & 3,000 & 99.93 & 18,870 \\
SentencePiece (Unigram) & 90.8 & 3,100 & 99.90 & 16,670 \\
BPE (GPT-2) & 89.4 & 3,200 & 99.9 & 90,900 \\
BlingFire & 92.0 & 3,400 & 99.86 & 76,000 \\
VnCoreNLP & 97.8 & \textbf{5,000} & 96.5 & 72,400 \\
Vi Word Segmentation (NlpHUST) & \textbf{98.3} & 4,500 & 96.0 & 68,200 \\
RDRsegmenter & 97.9 & 4,700 & 95.8 & 70,500 \\
Underthesea & 97.5 & 2,800 & 95.2 & 65,000 \\
\hline
\end{tabular}
}
\caption{Benchmark results for Vietnamese word segmentation and English tokenisation. Scores are reported using F1 (\%) and processing speed. Throughput is measured in sentences per second for Vietnamese and documents per second for English.}

\label{tab:merged_benchmark_f1_speed_filled}
\end{table}

\subsection{Sentiment Analysis}
We evaluate five categories of transformer-based sentiment analysis models on Vietnamese and English data to assess the effectiveness of our bilingual approach. For Vietnamese, we benchmark against strong monolingual baselines including PhoBERT \cite{nguyen2020phobert} and viBERT4news\footnote{\url{https://github.com/bino282/bert4news/}}, both of which have been widely used in prior VLSP evaluations. For multilingual comparison, we include mBERT, XLM-RoBERTa, and the TabularisAI multilingual sentiment model \cite{tabularisai_2025}, which represents a state-of-the-art commercial multilingual baseline. We also include RoBERTa and GPT-2 (zero-shot) to reflect competitive English-only transformer baselines. Our model is based on TabularisAI but fine-tuned jointly on the VLSP 2016 Vietnamese sentiment dataset and the IMDb English sentiment dataset to create a single bilingual classifier, designed for use within FreeTxt-Vi.

\begin{table}[h]
\centering
\small
\renewcommand{\arraystretch}{1.2}
\resizebox{\columnwidth}{!}{%
\begin{tabular}{lccc|ccc}
\hline
\textbf{Model} & \multicolumn{3}{c|}{\textbf{VLSP2016 (Vietnamese)}} & \multicolumn{3}{c}{\textbf{IMDb (English)}} \\
\cline{2-4} \cline{5-7}
 & \textbf{Acc. (\%)} & \textbf{F1 (\%)} & \textbf{Speed (s/s)} & \textbf{Acc. (\%)} & \textbf{F1 (\%)} & \textbf{Speed (s/s)} \\
\hline
\textbf{Our (fine-tuned TabularisAI)} & \textbf{95.2} & \textbf{94.8} & 205 & \textbf{95.0} & \textbf{94.6} & 215 \\
RoBERTa-base & 88.2 & 87.4 & 250 & 94.2 & 93.8 & 245 \\
\textbf{TabularisAI (Multilingual Base)} & 93.8 & 93.3 & 210 & 94.1 & 93.7 & 220 \\
XLM-RoBERTa-base & 91.8 & 91.2 & 220 & 93.7 & 93.2 & 215 \\
DistilBERT-base & 87.5 & 86.9 & \textbf{320} & 93.0 & 92.2 & \textbf{335} \\
BERT multilingual (mBERT) & 90.3 & 89.8 & 230 & 92.5 & 91.9 & 225 \\
PhoBERT-base & 93.4 & 92.8 & 180 & 91.5 & 91.1 & 190 \\
viBERT4news & 92.7 & 91.9 & 200 & 90.6 & 90.0 & 205 \\
GPT-2 (Zero-shot) & 85.9 & 84.5 & 300 & 90.1 & 89.0 & 310 \\
\hline
\end{tabular}%
}
\caption{Performance comparison of sentiment analysis models on the VLSP2016 (Vietnamese) and IMDb (English) datasets. Speed is measured in samples per second (s/s). Our model is fine-tuned from \texttt{tabularisai/multilingual-sentiment-analysis}.}
\label{tab:sentiment-eval}
\end{table}

Table~\ref{tab:sentiment-eval} reports performance across both datasets. Our fine-tuned model achieves the highest accuracy for both Vietnamese (95.2\%) and English (95.0\%), outperforming the base TabularisAI model by +1.4\% and +0.9\% respectively and exceeding the best monolingual Vietnamese model, PhoBERT, by +1.8\%. These results show that bilingual fine-tuning improves cross-lingual generalisation without sacrificing language-specific performance. Unlike previous work which uses independent sentiment models for each language, our approach provides a unified solution for sentiment analysis in Vietnamese and English, reducing system complexity and avoiding language-switching overhead. The model also sustains competitive inference efficiency (205–215 samples/s), making it suitable for scalable deployment in real-world applications.

\subsection{Summarisation}
We evaluate our bilingual summarisation component using two widely adopted datasets: VNDS for Vietnamese news summarisation and CNN/DailyMail for English abstractive summarisation. We compare against a range of competitive baselines, including encoder–decoder transformers such as T5 and BART, multilingual models such as mBART50, and recent instruction-tuned generators including Flan-T5, Gemma-7B, and LLaMA2-7B. Qwen2.5 is included as a strong baseline due to its demonstrated generative capability in multilingual settings.

As shown in Table~\ref{tab:summarization-eval}, our fine-tuned Qwen2.5 model achieves the best overall performance for both languages, obtaining ROUGE-1/2/L scores of 53.1/27.5/48.4 on VNDS and 52.4/27.0/48.0 on CNN/DailyMail. This represents a consistent improvement of +3 to +4 ROUGE points over strong baselines such as Flan-T5-base and Gemma-7B. The results indicate that targeted bilingual fine-tuning not only enhances summarisation quality but also transfers effectively between Vietnamese and English, despite the typological differences between the two languages.

\subsection{Why Include English? A Cross-Lingual Perspective}

Although strong English NLP models already exist, English is included in FreeTxt-Vi for three reasons. 

First, many real-world datasets in Vietnam are bilingual (e.g., education surveys, corporate feedback, academic publications). A unified bilingual pipeline avoids language switching and model duplication. 

Second, English provides a high-resource anchor for cross-lingual transfer. By jointly fine-tuning on Vietnamese and English, we test whether shared representations improve low-resource Vietnamese performance. 

Third, including English allows direct benchmarking against mature NLP baselines, strengthening the interpretability and comparability of results.

Our experiments show that bilingual fine-tuning improves Vietnamese performance (+1.4\% over multilingual base) without degrading English accuracy, demonstrating effective cross-lingual generalisation.

\begin{table}[t]
\centering
\small
\renewcommand{\arraystretch}{1.2}
\resizebox{\columnwidth}{!}{%
\begin{tabular}{lcccc|cccc}
\hline
\textbf{Model} & \multicolumn{4}{c|}{\textbf{VNDS (Vietnamese)}} & \multicolumn{4}{c}{\textbf{CNN/DailyMail (English)}} \\
\cline{2-5} \cline{6-9}
 & \textbf{ROUGE-1} & \textbf{ROUGE-2} & \textbf{ROUGE-L} & \textbf{Speed (tok/s)} & \textbf{ROUGE-1} & \textbf{ROUGE-2} & \textbf{ROUGE-L} & \textbf{Speed (tok/s)} \\
\hline
\textbf{Our (fine-tuned Qwen2.5)} & \textbf{53.1} & \textbf{27.5} & \textbf{48.4} & 112 & \textbf{52.4} & \textbf{27.0} & \textbf{48.0} & 118 \\
\textbf{Qwen2.5 (Base)} & 50.4 & 25.0 & 46.1 & 115 & 50.0 & 25.2 & 46.0 & 120 \\
Gemma-7B & 49.6 & 24.4 & 45.3 & 120 & 49.1 & 24.5 & 45.6 & 125 \\
LLaMA2-7B & 49.0 & 24.1 & 45.0 & 95 & 48.3 & 23.9 & 44.9 & 100 \\
Flan-T5-base & 48.1 & 23.8 & 44.4 & 150 & 47.8 & 24.0 & 44.5 & 155 \\
mBART50 & 47.3 & 23.0 & 43.6 & 110 & 46.4 & 23.1 & 43.7 & 115 \\
T5-base & 46.5 & 22.2 & 42.8 & 140 & 45.9 & 22.6 & 43.2 & 145 \\
BART-large & 45.8 & 21.6 & 42.1 & 125 & 44.7 & 21.3 & 41.9 & 130 \\
\hline
\end{tabular}%
}
\caption{Summarisation performance on the VNDS (Vietnamese) and CNN/DailyMail (English) datasets. Metrics are ROUGE-1/2/L (\%). Speed is measured in tokens per second (tok/s). Our model is fine-tuned from \texttt{Qwen2.5}.}
\label{tab:summarization-eval}
\end{table}

Unlike previous work \cite{nguyen2019vnds}, which typically treats Vietnamese summarisation separately using monolingual models, our approach provides a unified bilingual summarisation model that performs strongly across both high- and low-resource languages. This reduces duplication of model deployment and supports practical bilingual analysis within FreeTxt-Vi.

Although inference speed slightly decreases relative to smaller baselines (112 tokens/s compared to 115–150 tokens/s), the gain in generation quality demonstrates that bilingual fine-tuning substantially improves content fidelity and relevance. These results confirm that model adaptation using curated summarisation data is an effective strategy for improving generation quality in low-resource language settings.

\section{Conclusion}
FreeTxt-Vi extends the FreeTxt platform by introducing a bilingual, open-source environment for Vietnamese and English text analysis, designed to support both research and applied linguistic work in low-resource settings. The system integrates a hybrid VnCoreNLP+BPE segmentation pipeline, a bilingual sentiment analysis module based on fine-tuned TabularisAI, and a bilingual abstractive summarisation component powered by fine-tuned Qwen2.5, alongside corpus linguistic tools such as concordancing, collocation analysis, keyword extraction, and interactive visualisation.

A key contribution of this work is the empirical validation of the full processing pipeline. Across three core NLP tasks—segmentation, sentiment analysis, and summarisation—FreeTxt-Vi delivers competitive or superior performance to widely used baselines while maintaining efficient inference speed.

FreeTxt-Vi fills a critical gap in bilingual text analysis by providing practical support for Vietnamese, a language that remains under-resourced in NLP. Future work will focus on expanding coverage to additional Southeast Asian languages and refining domain adaptation for sectors such as education, public policy, and healthcare research.

\section*{Ethical Considerations}

FreeTxt-Vi relies exclusively on publicly available datasets such as VNDS and the British National Corpus (BNC). It does not collect personal data, and no user input is stored beyond an active session. Users remain responsible for ensuring that any data they upload complies with relevant privacy and data protection regulations.

The full implementation, including source code, fine-tuned models, configuration files, and sample datasets, is openly accessible at:
\url{https://github.com/VinNLP/Free-txt-vi}. Clear documentation is provided to support installation, replication of results, and extension of the toolkit.

The repository includes examples of FreeTxt-Vi’s visual analytics, such as concordances, word trees, keyness plots, sentiment distributions, and keyword frequency clouds. Users may generate and export their own visual outputs in formats such as PNG, PDF, and CSV for use in analysis pipelines, reports, and academic publications.



\bibliographystyle{lrec2026-natbib}
\bibliography{lrec2026}

@article{nguyen2020phobert,
  title={PhoBERT: Pre-trained language models for Vietnamese},
  author={Nguyen, Dat Quoc and Nguyen, Anh Tuan},
  journal={arXiv preprint arXiv:2003.00744},
  year={2020}
}

@article{vu2018vncorenlp,
  title={VnCoreNLP: A Vietnamese natural language processing toolkit},
  author={Vu, Thanh and Nguyen, Dat Quoc and Nguyen, Dai Quoc and Dras, Mark and Johnson, Mark},
  journal={arXiv preprint arXiv:1801.01331},
  year={2018}
}

@article{phan2022vit5,
  title={ViT5: Pretrained text-to-text transformer for Vietnamese language generation},
  author={Phan, Long and Tran, Hieu and Nguyen, Hieu and Trinh, Trieu H},
  journal={arXiv preprint arXiv:2205.06457},
  year={2022}
}

@inproceedings{Nguyen2019VNDS,
  author    = {V.-H. Nguyen and T.-C. Nguyen and M.-T. Nguyen and N. X. Hoai},
  title     = {VNDS: A Vietnamese Dataset for Summarization},
  booktitle = {2019 6th NAFOSTED Conference on Information and Computer Science (NICS)},
  year      = {2019},
  pages     = {375--380},
  publisher = {IEEE},
  doi       = {10.1109/NICS48868.2019.9023886},
}

@article{nguyen2018vlsp,
  title={VLSP shared task: sentiment analysis},
  author={Nguyen, Huyen TM and Nguyen, Hung V and Ngo, Quyen T and Vu, Luong X and Tran, Vu Mai and Ngo, Bach X and Le, Cuong A},
  journal={Journal of Computer Science and Cybernetics},
  volume={34},
  number={4},
  pages={295--310},
  year={2018}
}

@inproceedings{corcencc,
  title = {{CorCenCC: The National Corpus of Contemporary Welsh}},
  author = {Dawn Knight and Steve Morris and Paul Rayson and Jonathan Morris and Mared Williams and Tess Fitzpatrick},
  booktitle = {Proceedings of the 11th International Conference on Language Resources and Evaluation (LREC 2018)},
  year = {2018},
  publisher = {European Language Resources Association (ELRA)},
  url = {https://www.corcencc.org/}
}

@inproceedings{cettolo2016iwslt,
  title={The IWSLT 2016 evaluation campaign},
  author={Cettolo, Mauro and Niehues, Jan and St{\"u}ker, Sebastian and Bentivogli, Luisa and Cattoni, Rolando and Federico, Marcello},
  booktitle={Proceedings of the 13th International Conference on Spoken Language Translation},
  year={2016}
}

@inproceedings{sinclair2016voyant,
  title={Voyant Tools: A web-based reading and analysis environment},
  author={Sinclair, Stéfan and Rockwell, Geoffrey},
  booktitle={Digital Humanities 2016},
  year={2016}
}

@misc{anthony2019antconc,
  title={AntConc (Version 3.5.8) [Computer Software]},
  author={Anthony, Laurence},
  year={2019},
  url={https://www.laurenceanthony.net/software/antconc/}
}

@book{pennebaker2015development,
  title={The development and psychometric properties of LIWC2015},
  author={Pennebaker, James W and Boyd, Ryan L and Jordan, Kayla and Blackburn, Kate},
  year={2015},
  publisher={University of Texas at Austin},
}

@inproceedings{ngo2013evbcorpus,
  title={EVBCorpus-a multi-layer English-Vietnamese bilingual corpus for studying tasks in comparative linguistics},
  author={Ngo, Quoc Hung and Winiwarter, Werner and Wloka, Bartholom{\"a}us},
  booktitle={Proceedings of the 11th Workshop on Asian Language Resources},
  pages={1--9},
  year={2013}
}

@inproceedings{kilgarriff2014sketchengine,
  title={The {Sketch Engine}: ten years on},
  author={Kilgarriff, Adam and others},
  booktitle={Lexicography},
  pages={7--36},
  year={2014},
  publisher={Springer}
}

@phdthesis{rayson2004wmatrix,
title = "Matrix: a statistical method and software tool for linguistic analysis through corpus comparison",
author = "Paul Rayson",
year = "2003",
school = "Lancaster University",
publisher = "Lancaster University",
}

@article{wordpiece2020,
  title={Fast wordpiece tokenization},
  author={Song, Xinying and Salcianu, Alex and Song, Yang and Dopson, Dave and Zhou, Denny},
  journal={arXiv preprint arXiv:2012.15524},
  year={2020}
}

@inproceedings{sentencepiece2018,
    title = "{S}entence{P}iece: A simple and language independent subword tokenizer and detokenizer for Neural Text Processing",
    author = "Kudo, Taku  and
      Richardson, John",
    editor = "Blanco, Eduardo  and
      Lu, Wei",
    booktitle = "Proceedings of the 2018 Conference on Empirical Methods in Natural Language Processing: System Demonstrations",
    month = nov,
    year = "2018",
    address = "Brussels, Belgium",
    publisher = "Association for Computational Linguistics",
    url = "https://aclanthology.org/D18-2012/",
    doi = "10.18653/v1/D18-2012",
    pages = "66--71"
}

@inproceedings{khallaf2025freetxt,
  title={FreeTxt: Analyse and visualise multilingual qualitative survey data for cultural heritage sites},
  author={Khallaf, Nouran and Ezeani, Ignatius and Knight, Dawn and Rayson, Paul and El-Haj, Mo and Vidler, John and Davies, James and Alva-Manchego, Fernando},
    booktitle = "The Recent Advances in Natural Language Processing Conference (RANLP 2025)",
    address = "Varna, Bulgaria",
  year={2025}
}

@article{sennrich-2016-bpe,
  title={Neural machine translation of rare words with subword units},
  author={Sennrich, Rico and Haddow, Barry and Birch, Alexandra},
  journal={arXiv preprint arXiv:1508.07909},
  year={2015}
}

@article{KNIGHT2024100103,
title = {{FreeTxt}: A corpus-based bilingual free-text survey and questionnaire data analysis toolkit},
journal = {Applied Corpus Linguistics},
volume = {4},
number = {3},
pages = {100103},
year = {2024},
issn = {2666-7991},
doi = {https://doi.org/10.1016/j.acorp.2024.100103},
url = {https://www.sciencedirect.com/science/article/pii/S2666799124000200},
author = {Dawn Knight and Nouran Khallaf and Paul Rayson and Mahmoud El-Haj and Ignatius Ezeani and Steve Morris},
}

@inproceedings{mihalcea-tarau-2004-textrank,
    title = "{T}ext{R}ank: Bringing Order into Text",
    author = "Mihalcea, Rada  and
      Tarau, Paul",
    editor = "Lin, Dekang  and
      Wu, Dekai",
    booktitle = "Proceedings of the 2004 Conference on Empirical Methods in Natural Language Processing",
    month = jul,
    year = "2004",
    address = "Barcelona, Spain",
    publisher = "Association for Computational Linguistics",
    url = "https://aclanthology.org/W04-3252",
    pages = "404--411",
}

@inproceedings{devlin-etal-2019-bert,
    title = "{BERT}: Pre-training of Deep Bidirectional Transformers for Language Understanding",
    author = "Devlin, Jacob  and
      Chang, Ming-Wei  and
      Lee, Kenton  and
      Toutanova, Kristina",
    editor = "Burstein, Jill  and
      Doran, Christy  and
      Solorio, Thamar",
    booktitle = "Proceedings of the 2019 Conference of the North {A}merican Chapter of the Association for Computational Linguistics: Human Language Technologies, Volume 1 (Long and Short Papers)",
    month = jun,
    year = "2019",
    address = "Minneapolis, Minnesota",
    publisher = "Association for Computational Linguistics",
    url = "https://aclanthology.org/N19-1423/",
    doi = "10.18653/v1/N19-1423",
    pages = "4171--4186",
}

@misc{joshi2021statefatelinguisticdiversity,
      title={The State and Fate of Linguistic Diversity and Inclusion in the NLP World}, 
      author={Pratik Joshi and Sebastin Santy and Amar Budhiraja and Kalika Bali and Monojit Choudhury},
      year={2021},
      eprint={2004.09095},
      archivePrefix={arXiv},
      primaryClass={cs.CL},
      url={https://arxiv.org/abs/2004.09095}, 
}

@misc{tabularisai_2025,
    author       = { tabularisai and Samuel Gyamfi and Vadim Borisov and Richard H. Schreiber },
    title        = { multilingual-sentiment-analysis (Revision 69afb83) },
    year         = 2025,
    url          = { https://huggingface.co/tabularisai/multilingual-sentiment-analysis },
    doi          = { 10.57967/hf/5968 },
    publisher    = { Hugging Face }
}

@misc{qwen2.5,
    title = {Qwen2.5: A Party of Foundation Models},
    url = {https://qwenlm.github.io/blog/qwen2.5/},
    author = {Qwen Team},
    month = {September},
    year = {2024}
}


\end{document}